\documentclass[letterpaper,10 pt,conference]{ieeeconf}  

\usepackage{tabularx}

\IEEEoverridecommandlockouts                              
\usepackage{cite}
\usepackage{array}
\usepackage{graphics} 
\usepackage{subfig}
\usepackage{multirow}

\usepackage{url}
\usepackage{booktabs}
\usepackage[detect-all,per-mode=symbol]{siunitx}
\usepackage{dsfont}
\usepackage{algorithmicx}
\usepackage[noend]{algpseudocode}
\usepackage{graphicx,caption}
\usepackage{amsmath,amssymb,stmaryrd,mathtools}
\usepackage[keeplastbox]{flushend}
\usepackage{algorithm}
\usepackage{hyperref}
\usepackage{acronym}
\usepackage{color,xcolor}
\usepackage{verbatim}
\usepackage{pgfplots}
\usepackage{filecontents}
\usepackage{xcolor}
\usepackage{tikz}
\definecolor{bblue}{HTML}{4F81BD}
\definecolor{rred}{HTML}{C0504D}
\definecolor{ggreen}{HTML}{9BBB59}
\definecolor{ppurple}{HTML}{9F4C7C}
\pgfplotsset{compat=1.11,
    /pgfplots/ybar legend/.style={
    /pgfplots/legend image code/.code={%
       \draw[##1,/tikz/.cd,yshift=-0.25em]
        (0cm,0cm) rectangle (3pt,0.8em);},
   },
}
\usepackage{gensymb}
\usepackage{slantsc}
\newcommand{\norm}[1]{\left\lVert #1 \right\rVert} 

\graphicspath{{./figures/}}

\newcommand{\ignore}[1]{}

\usepackage{cleveref}
\newcommand{\citet}[1]{\citeauthor{#1}~\shortcite{#1}}

\DeclareFontFamily{U}{MnSymbolC}{}
\DeclareSymbolFont{MnSyC}{U}{MnSymbolC}{m}{n}
\DeclareFontShape{U}{MnSymbolC}{m}{n}{
    <-6>  MnSymbolC5
   <6-7>  MnSymbolC6
   <7-8>  MnSymbolC7
   <8-9>  MnSymbolC8
   <9-10> MnSymbolC9
  <10-12> MnSymbolC10
  <12->   MnSymbolC12%
}{}
\DeclareMathSymbol{\powerset}{\mathord}{MnSyC}{180}

\definecolor{orange}{rgb}{1, .36, .08}
\definecolor{darkmagenta}{rgb}{0.698,0,0.698}

\definecolor{vg_edit_color}{rgb}{0, 0.0, 1.0}

\usepackage{todonotes}
\usepackage{soul}
\definecolor{smoothgreen}{rgb}{0.7,1,0.7}
\sethlcolor{smoothgreen}

\renewcommand{\baselinestretch}{0.99}

\title{\LARGE \bf HG-DAgger: Interactive Imitation Learning with Human Experts}

\author{Michael Kelly, Chelsea Sidrane, Katherine Driggs-Campbell, and Mykel J. Kochenderfer
\thanks{
This material is based upon work supported by SAIC Innovation Center, a subsidiary of SAIC Motors and by AFRL and DARPA under contract FA8750-18-C-0099.}
\thanks{M. Kelly is with the Computer Science Department, Stanford University, Stanford, CA, USA (e-mail: mkelly2@stanford.edu). C. Sidrane and M.J. Kochenderfer are with the Aeronautics and Astronautics Department, Stanford University, Stanford, CA, USA (e-mail: \{csidrane,mykel\}@stanford.edu). K. Driggs-Campbell is with the Electrical and Computer Engineering Department, University of Illinois at Urbana-Champaign, Urbana, IL (e-mail: krdc@illinois.edu).}%
}

\begin{document}

\maketitle

\begin{abstract}
Imitation learning has proven to be useful for many real-world problems, but approaches such as behavioral cloning suffer from data mismatch and compounding error issues. One attempt to address these limitations is the \textsc{DAgger} algorithm, which uses the state distribution induced by the novice to sample corrective actions from the expert. Such sampling schemes, however, require the expert to provide action labels without being fully in control of the system. This can decrease safety and, when using humans as experts, is likely to degrade the quality of the collected labels due to perceived actuator lag. In this work, we propose HG-\textsc{DAgger}, a variant of \textsc{DAgger} that is more suitable for interactive imitation learning from human experts in real-world systems. In addition to training a novice policy, HG-\textsc{DAgger} also learns a safety threshold for a model-uncertainty-based risk metric that can be used to predict the performance of the fully trained novice in different regions of the state space. We evaluate our method on both a simulated and real-world  autonomous driving task, and demonstrate improved performance over both \textsc{DAgger} and behavioral cloning.
\end{abstract}


\section{Introduction}
\label{sec:related_work}

Imitation learning is often posed as a supervised learning problem. Data is gathered from an expert policy and is used to train a novice policy~\cite{price2003accelerating,Kober2010}. While this approach, known as behavioral cloning, can be an effective way to learn policies in scenarios where there is sufficiently broad data coverage, in practice it often suffers from data mismatch and compounding errors~\cite{ross2010efficient}.  

Online sampling frameworks such as the \textsc{DAgger} algorithm have been proposed to address the drawbacks inherent in naive behavioral cloning~\cite{ross2010efficient}. \textsc{DAgger} trains a sequence of novice policies using corrective action labels provided by the expert at states sampled by a mixture of the expert and the novice policies. At each time-step in a data-gathering rollout, a gating function determines which of the two policies' choice of action will actually be executed on the combined system; in the case of \textsc{DAgger}, this gating function amounts to a weighted coin toss that executes the expert's choice of action with some probability $\beta \in [0,1]$ and the novice's choice of action with probability $1 - \beta$. 

By allowing the novice to influence the sampling distribution used to acquire expert action labels (a practice known as ``Robot-Centric" (RC) sampling), \textsc{DAgger} trains a more robust policy that is capable of handling perturbations from the nominal trajectories of the expert~\cite{laskey2017comparing}. While \textsc{DAgger} has many appealing properties, including theoretical performance guarantees, its use of RC sampling can compromise training-time safety because it involves performing data-gathering rollouts under the partial or complete control of an incompletely trained novice. Furthermore, the shared control scheme can alter the behavior of human experts, degrading the quality of the sampled action labels and potentially even destabilizing the combined system.

The central problem with RC sampling methods is that they may not provide the human expert with sufficient control authority during the sampling process~\cite{laskey2017comparing}. Humans rely on feedback from the system under control to perform optimally; however, many RC sampling based methods (for example, \textsc{DAgger} with $\beta=0$) provide the human expert with no feedback signal at all, since the human's choice of action is never actually executed.

RC sampling methods that do allow the expert to influence the sampling process, such as \textsc{DAgger} with nonzero $\beta$, switch control authority back and forth between the novice and the expert during training, which can degrade the feedback signal to the user and cause a perceived actuator lag. 
Humans tend to be sensitive to small changes in execution, like time delay~\cite{goil2013using,kim1992force}. 
These issues can drive the human expert to adapt and change their behaviors over time~\cite{shia2014semiautonomous}. 
In imitation learning, this can degrade learning stability and may ultimately cause the novice to learn behaviors that are significantly different from the unbiased expert behavior~\cite{kim1992force}. 


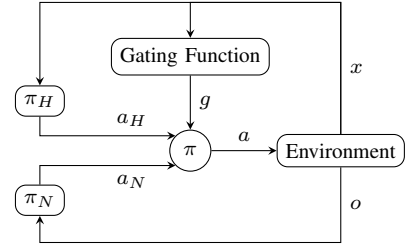
\begin{figure}[!t]
    \centering
    \usetikzlibrary{shapes.geometric, arrows}
\tikzstyle{startstop} = [rectangle, rounded corners, minimum width=1.5cm, minimum height=.75cm,text centered, draw=black, fill=none]
\tikzstyle{arrow}=[draw]
\tikzset{%
    arrow/.style={->,>=stealth},
}

\begin{tikzpicture}[node distance=0.9cm,font=\footnotesize]

\tikzstyle{block} = [rectangle, rounded corners,text centered, draw=black, fill=none, node distance=1cm]
\tikzstyle{round} = [draw=black, circle, fill=none, node distance=1cm, text centered]
\tikzstyle{invisible} = [fill=none, inner sep=0pt]
\tikzstyle{line} = [draw, -stealth]
    
    \node [block] (gf) {Gating Function};
    \node [round, below of=gf, yshift=-.25cm] (pi) {$\pi$};
    \node [block, right of=pi, xshift=1cm] (env) {Environment};
    \node [block, left of=pi, yshift=.66cm, xshift=-1cm] (pihuman) {$\pi_H$};
    \node [block, left of=pi, yshift= -.66cm, xshift=-1cm] (pinovice) {$\pi_N$};
    \node [invisible, above of=gf, yshift=.1cm] (invis) {};

    \path [line] (pi) -- node [anchor=south] {$a$} (env);
    \path [line] (pihuman) |- node [near end, anchor=south west] {$a_H$} (pi.north west);
    \path [line] (pinovice) |- node [near end, anchor=north west] {$a_N$} (pi.south west);
    \path [line] (gf.south) -- node [anchor=west] {$g$} (pi);
    \path [line] (env.north) |- node [near start, anchor=west] {$x$} ++(-2,+1.75) -- (gf.north);
    \path [line] (env.north) |- ++(-2,+1.75) -| (pihuman); 
    \path [line] (env.south) |- node [near start, anchor=west] {$o$} ++(-4,-1) -- (pinovice.south);

\end{tikzpicture}

    \caption{\small Control loop for HG-\textsc{DAgger}} 
    \label{fig:flowchart}
    \vspace{-10pt} 
\end{figure}

Furthermore, even for two policies that are individually stable, there may exist some switching procedures that destabilize the combined system~\cite{branicky1998multiple}. 
For example,
in the aeronautical domain, pilot-induced oscillations can arise from a series of (over) corrections from a pilot attempting to stabilize an aircraft~\cite{mcruer1995pilot}.
When shared control systems directly affect the pilot's inputs, they are often aware that their authority over the control of the system is reduced.  
This awareness often leads to a greater response (or an overcorrection) from the pilot, in turn causing conflicting efforts from the control system and the human~\cite{national1997aviation}.


In the case of the \textsc{DAgger} algorithm, tuning the $\beta$ parameter schedule may mitigate these issues. A larger $\beta$-value that is reduced more slowly over training epochs will provide greater control authority to the human expert, which should allow for the collection of higher-quality action labels and improved safety during training, since the potentially unsafe novice has a lesser degree of control. However, increasing $\beta$ too much will likely degrade test-time performance, particularly in the extreme case of $\beta=1$, where \textsc{DAgger} reduces to behavioral cloning and is subject to compounding error issues. It is therefore not clear how or if a good choice of $\beta$ for an arbitrary imitation learning task can be made \textit{a priori}, and tuning $\beta$ on a physical system is likely to be challenging due both to the expense of collecting samples from real-world systems and due to $\beta$'s effect on training-time safety~\cite{kim1992force}. Furthermore, it is not necessarily clear that there exists a $\beta$ that provides a satisfactory trade-off between learning performance and training-time safety.

Recent prior work using human experts has not fully addressed these issues, and has focused instead primarily on reducing the work load of the expert by minimizing the number of times the expert is queried~\cite{zhang2016query,laskey2016shiv,kim2013maximum}.  
Instead of querying at every time step, the expert only gives corrective actions when there is a significant discrepancy between the anticipated behaviors of the novice and the expert. 
These approaches effectively reduce the number of demonstrations and expert queries required while preserving strong theoretical guarantees on performance and often improving safety.

In this work, we consider a probabilistic extension of \textsc{DAgger} called \textsc{EnsembleDAgger}.
This method takes a Bayesian approach to determine whether it is safe for the novice to act by approximating risk with familiarity~\cite{menda2018ensembledagger}.  
To do this, the novice is represented as an ensemble of neural networks; such ensembles can efficiently approximate Gaussian processes, which scale poorly with the amount of training data. Using the mean and variance of the policy's actions, the \emph{confidence} of the novice can be assessed and used to determine whether or not the expert should intervene.  
In short, this method aims to maximize the novice's share of actions during data-gathering rollouts, while constraining the probability of failure in a model-free manner.  
However, identifying measures of risk and appropriate thresholds on the variance remains an open problem in both model-free and model-based communities~\cite{menda2018ensembledagger,driggs2015improved,govindarajan2017data}.

We expand upon the concepts proposed by \textsc{EnsembleDAgger} and develop a methodology for learning from humans in scenarios where running a novice policy in parallel with a human policy is not intuitive for the user.  
We present the following contributions:

\begin{enumerate}
    \item We propose Human-Gated \textsc{DAgger} (HG-\textsc{DAgger}), a \textsc{DAgger}-variant designed for more effective imitation learning from human experts.
    \item We propose a data-driven approach for learning a safety threshold for our accompanying risk metric and show that the metric produced is meaningful.
    \item We demonstrate the efficacy of our method on an autonomous driving task, showing improved sample efficiency, greater training stability, and more human-like behavior relative to $\textsc{DAgger}$ and behavioral cloning.
\end{enumerate}

This paper is organized as follows.
Section~\ref{sec:methods} presents the methodology for our proposed framework, HG-\textsc{DAgger}.
The experimental setup for collecting human data and training novice policies is presented in Section~\ref{sec:experiments}.
The performance of the resulting policies is analyzed in Section~\ref{sec:results}.
Section~\ref{sec:disc} discusses our findings and outlines future work.
\section{Methods}
\label{sec:methods}


We are motivated by the assumption that higher quality action labels can be acquired when the human expert is given stretches of uninterrupted control authority. Other methods hand off control stochastically at each time-step, or require that the human expert retroactively provide corrective action labels to states visited by the novice. Instead, we allow the human expert to take control when they deem it necessary and to maintain exclusive control authority until they manually hand control back to the novice policy.

Using nomenclature borrowed from the hierarchical RL literature, we refer to this human supervisor as a gating function, which decides whether the expert or novice ``sub-policy" should be in control at a given moment. For this reason, we call our algorithm Human-Gated \textsc{DAgger} (HG-\textsc{DAgger}). Fig.~\ref{fig:flowchart} illustrates this control scheme.


Like other \textsc{DAgger} variants, HG-\textsc{DAgger} trains a sequence of novice policies on a training data set $\mathcal{D}$, which is iteratively augmented with additional expert labels collected during repeated data-gathering rollouts of a combined expert-novice system. Unlike other \textsc{DAgger} variants, however, the gating function employed by HG-\textsc{DAgger} is controlled directly by the expert. In HG-\textsc{DAgger}, the novice policy is rolled out until the expert observes that the novice has entered an unsafe region of the state space. 
The expert takes control and guides the system back to a safe and stable region of the state space. Expert action labels are only collected and added to $\mathcal{D}$ during these recovery trajectories, during which the human expert has uninterrupted control of the system. Once in a safe region, control is returned to the novice.

We refer to the set of ``safe" states (as judged by the human expert) as the \emph{permitted set} $x_t \in \mathcal{P}$ and formalize the human-controlled gating function accordingly as 
$g(x_t) = \mathds{1}[x_t \not\in \mathcal{P}]$ 
Given a human expert $\pi_H$ and the current instantiation of the novice $\pi_{N_i}$, we can then express the \textsc{HG-DAgger} rollout policy for the $i$th training epoch, $\pi_{i}$, as:
\begin{align}
   \pi_{i}(x_t)
   &=
   g(x_t) \pi_H(x_t) + (1 - g(x_t)) \pi_{N_i}(o_t)
\end{align}
where $o_t$ is the observation received by the novice at the current state $x_t$, generated by the observation function $\mathcal{O}(\cdot)$. A human expert has access to many channels of information that are unavailable to the novice policy; we acknowledge this by explicitly denoting that the expert has access to the full state, $x_t$, but that the novice only has access to an observation, $o_t=\mathcal{O}(x_t)$.
Letting $\xi_i$ represent the concatenation of all rollouts performed with $\pi_{i}$, we can represent the data collected in the $i$th training epoch as
\begin{equation}
    \mathcal{D}_i
    =
    \{ (\mathcal{O}(x_t),\pi_H(x_t))
    \mid
    g(x_t) = 1,
    x_t \in \xi_i
    \}
\end{equation}
At the end of epoch $i$, $\mathcal{D}_i$ is added to the training data set $\mathcal{D}$ and the next novice policy is trained on the aggregated data. The dataset is initialized with a set of samples $\mathcal{D}_{BC}$ gathered using behavioral cloning.


In theory, the data collected with HG-\textsc{DAgger} is used to teach the novice to stabilize itself about the nominal expert trajectories demonstrated by the expert during the initial behavioral cloning step. Since both behavioral cloning and HG-\textsc{DAgger} collect data from the human expert only while it has uninterrupted control, we can expect to acquire high-quality action labels both along nominal expert trajectories and along recovery trajectories using this method.

At training time, HG-\textsc{DAgger} relies on the human expert to ensure safety by intervening in dangerous situations.\footnote{As a result, HG-\textsc{DAgger} is not suitable for application in those real-world domains where the human expert cannot quickly identify and react to unsafe situations.} At test time, the policy trained with these additional demonstrations is allowed to act without any human intervention. Meanwhile, HG-\textsc{DAgger} also learns a risk metric derived from the novice policy's ``doubt" that can be used to understand and assess the performance of the final trained policy. 

We use a risk approximation method inspired by~\cite{menda2018ensembledagger} and represent the novice as an ensemble of neural networks. 
The covariance matrix, $C_t$, of the ensemble's outputs given input $o_t$ contains useful measures of policy confidence. We use the $\ell_2$-norm of the main diagonal of $C_t$ as the doubt, $d_N(o_t)$.
\begin{equation}
    d_N(o_t) = \norm{\text{diag}(C_t)}_2
\end{equation}
Following the assumption that a neural network ensemble approximates a Gaussian process~\cite{lakshminarayanan2017simple}, we expect that doubt will be high in regions of the state space that are poorly represented in the data set. The novice is expected to perform poorly in these regions both because they are inadequately sampled and because they are likely to be more intrinsically risky, as the expert is likely to bias sampling away from more intrinsically risky regions when possible. \textsc{EnsembleDAgger} uses this risk heuristic to improve training-time safety by designing a gating function that only permits the novice to act when its doubt falls below some threshold, but the authors do not provide a means of selecting that threshold. 

Rather than attempting to select a ``safe" threshold value for doubt \textit{a priori}, we instead learn a threshold, $\tau$, from human data. We record the novice's doubt at the point of human intervention in a doubt intervention logfile, $\mathcal{I}$. We compute $\tau$ as the mean of the final 25\% of the entries in $\mathcal{I}$: 
 \begin{equation}
     \tau = \frac{1}{\text{len}(\mathcal{I})/4}\sum_{i=\lfloor .75 N\rfloor}^{N}(\mathcal{I}[i])
 \end{equation}
 
We chose this to balance learning $\tau$ from a larger number of human interventions and learning $\tau$ from only the most relevant human interventions. The most relevant interventions are those made during rollouts of novice policies trained on more data. These rollouts are made with a policy which more closely resemble the fully trained policy.
 
\begin{algorithm}[t!]
\caption{HG-\textsc{DAgger}}\label{algo:hg_dagger}
    \begin{algorithmic}[1]
      \Procedure{HG-\textsc{DAgger}}{$\pi_H,\pi_{N_1},\mathcal{D}_{BC}$}
      \State $\mathcal{D} \gets \mathcal{D}_{BC}$
      \State $\mathcal{I} \gets [ ]$
      	\For{epoch $i = 1 : K$}
        	 \For{rollout $j = 1 : M$}
        	    \For{timestep $t \in T$ of rollout j} 
        	        \If{expert has control}
            	        \State record expert labels into $\mathcal{D}_j$
            	    \EndIf
        	        \If{expert is taking control}
            	        \State record doubt into $I_j$
                    \EndIf 
                \EndFor
        	    \State $\mathcal{D} \gets \mathcal{D} \cup \mathcal{D}_j$
        	    \State append $\mathcal{I}_j$ to $\mathcal{I}$
        	\EndFor
        \State train $\pi_{N_{i+1}}$ on $\mathcal{D}$	
    	\EndFor
    \State $\tau \gets f(\mathcal{I})$
    \State\Return $\pi_{N_{K+1}}, \tau$
    \EndProcedure
    \end{algorithmic}
\end{algorithm} 

The current work uses this threshold to evaluate and understand the fully-trained policy's performance, but we propose in future work to use this risk metric at test time to determine when control of the system should be returned from the trained policy to a safer, more conservative controller. See Algorithm~\ref{algo:hg_dagger} for a summary of the \textsc{HG-DAgger} algorithm.
 

The approach most similar to ours is the Confidence-Based Autonomy algorithm proposed in~\cite{chernova2009}. Like $\textsc{HG-DAgger}$, this algorithm allows the expert to provide corrective demonstrations to the novice whenever the expert deems it necessary. However, the method also allows the novice to request demonstrations from the expert when a measure of its confidence falls below some threshold, allowing for a switching behavior similar to that seen in $\textsc{DAgger}$, which we have argued is problematic in the human-in-the-loop setting. Our approach also differs from~\cite{chernova2009} in that we use an ensemble of neural network learners instead of Gaussian mixture models. This distinction is significant because having a sufficiently expressive learner is important if we are to adequately mimic expert behavior on complex tasks~\cite{heheCoaching, laskey2017comparing}. Another important difference between the two works is that $\textsc{HG-DAgger}$ learns the doubt threshold from the expert (via training-time interventions) rather than calculating it in an ad-hoc manner (i.e. by setting it as three times the average nearest neighbor distance in the dataset). 



\section{Experimental Setup}
\label{sec:experiments}

We apply our method to a complex, real-world task: learning autonomous driving policies from human drivers. Early autonomous driving researchers in the 1980s applied behavioral cloning to the lane keeping task~\cite{pomerleau1989alvinn}. More recently, end-to-end learning approaches have made use of both behavioral cloning and \textsc{DAgger} frameworks~\cite{bojarski2016end,pan2017agile}.
This work specifically targets human-in-the-loop learning, which, as previously discussed, is typically a difficult problem.

In this section, we present the experimental setup used to collect data and train our policy  both in simulation and on a physical test vehicle. We compare the performance of HG-\textsc{DAgger} to that of \textsc{DAgger} and behavioral cloning.

\subsection{Experimental Task}
The driving task involves an ego vehicle moving along a two-lane, single direction roadway populated with stationary cars. The ego vehicle must weave between the cars safely, without leaving the road. We perform this experiment both in simulation and on a real automobile.

In training, the obstacle cars are initialized on the roadway at 30 meter intervals. The spacing of the obstacles is randomized by $\pm$ 5 meters and the lane that the obstacle appears in (left or right) is randomized as well. The dimensions of the vehicles are \SI{4}{\meter}$\times$\SI{1.5}{\meter}, and each lane is \SI{3}{\meter} across.

The novice policy receives an observation of the ego vehicle's state consisting of distance from the median $y$, orientation $\theta$, speed $s$, distances to each edge of the current lane  $(l_l,l_r)$, and distances to the nearest leading obstacle in each lane $(d_l,d_r)$. The policy then issues steering angle and speed commands to the ego vehicle.

Performance is evaluated primarily on road departure and collision rates. A road departure is when the center of mass of the ego vehicle leaves the road while under control of the novice policy. Rates were calculated on a per meter basis.

\subsubsection{Training}

For both the simulated and real-world experiments, we first trained a policy using behavioral cloning on 10,000 action labels collected from the expert. This policy was employed as an initialization for the three methods tested.\footnote{A behavioral cloning initialization was used because both \textsc{DAgger} and HG-\textsc{DAgger} rely on the novice to shape the sampling distribution used to acquire action labels from the expert. The state distribution induced by a completely untrained novice policy places density on regions of the state space that will not be visited once the task is learned.} 
Each method then refined that initial policy over an additional five training epochs, each of which involved the accumulation of an additional 2,000 expert labels.\footnote{Except for the final HG-\textsc{DAgger} training epoch, which incorporated fewer expert labels due to the novice policy's ability to successfully avoid the unsafe regions of the state space without human intervention.} For \textsc{DAgger}, we initialized the $\beta$ parameter at 0.85 and decayed it by a factor of 0.85 at the end of each training epoch.

\subsubsection{Vehicle Experiments}
\begin{figure}[!t]
    \centering
    \includegraphics[width=\columnwidth]{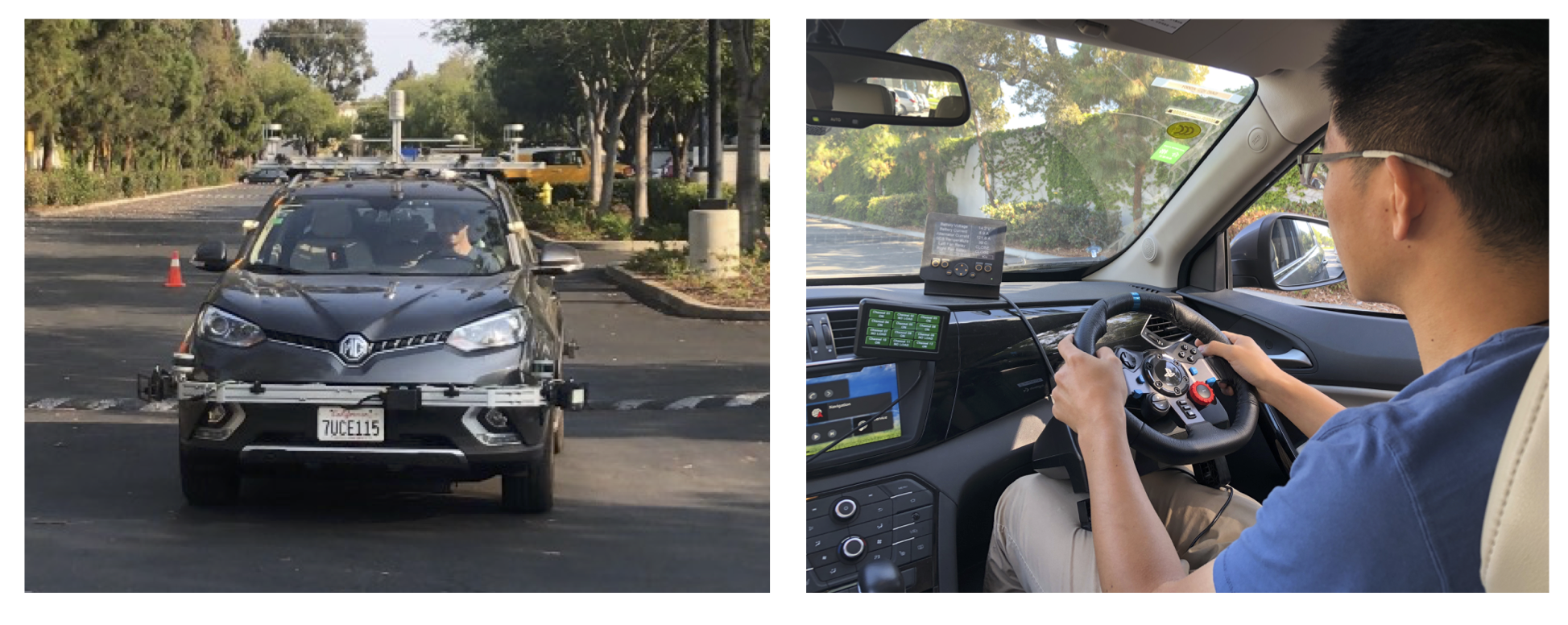}
    \caption{\small Test vehicle (L) and expert driver interface (R).}
    \label{fig:vehicle_setup}
    \vspace{-10pt}
\end{figure}

We trained and tested policies with each method 
using an MG-GS vehicle provided by SAIC as the ego vehicle, and simulated vehicles as the stationary obstacles. The MG-GS vehicle was equipped with LiDAR and high-fidelity localization. A two-lane road with static obstacle cars was simulated for testing. A safety driver monitored the vehicle at all times during testing and training, while another human driver who could see the simulated obstacles and road used an auxiliary steering wheel and pedal set to send control inputs to the system. For HG-\textsc{DAgger} data collection, the expert could retake control of the system by turning the steering wheel and could revert control to the novice by pressing a button. The vehicle and testing setup are shown in Figure~\ref{fig:vehicle_setup}.



\subsubsection{Simulation}
We trained additional policies in simulation in order to perform various evaluations that would be prohibitively time-consuming or dangerous to perform on a physical car. Ego vehicle dynamics were approximated using a bicycle model with parameters such as distance from center of gravity to axles determined by the physical test vehicle.\footnote{Simulations were implemented using AutomotiveDrivingModels.jl}

\section{Results}
\label{sec:results}

Our results demonstrate that novice policies trained with HG-\textsc{DAgger} outperform novice policies trained with \textsc{DAgger} and behavioral cloning in terms of sample efficiency, training stability, and similarity to human behavior in our driving task. Additionally, our results demonstrate the significance of the doubt threshold learned by our method.

\subsection{Simulation: Learning Performance}

We evaluate the effectiveness of each of the three methods as a function of the number of expert labels on which each novice policy was trained. 
The same eight randomly selected obstacle configurations and initializations were used for the evaluation of each policy. Learning curves from these tests are displayed in Figure~\ref{fig:lc_road_departs} and Figure~\ref{fig:lc_collision_rates}. The x-axis in each chart shows the number of expert labels on which each intermediate novice policy is trained.

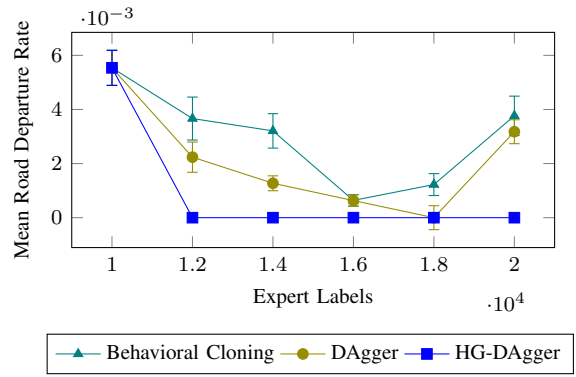
\begin{figure}[t!]
    \centering
    \begin{tikzpicture}[trim axis right]
    \begin{axis}[
    width=0.45*\textwidth,
    height=0.25*\textwidth,
    xlabel=Expert Labels,
    ylabel=Mean Road Departure Rate,
    font=\footnotesize,
    legend style={
    legend columns = -1,
    at={(1.04,-0.4)},
    font=\footnotesize}
    ]
        \addplot+[teal] [
                mark=triangle*,
                mark options={teal},
            error bars/.cd,
                y explicit,
                y dir=both,
        ] table [
            x index=0,
            y index=1,
            y error plus expr=\thisrow{BCSE},
            y error minus expr=\thisrow{BCSE},
        ] {departs_per_meter.txt};
        
        \addplot+[olive] [
                mark=*,
                mark options={olive},
            error bars/.cd,
                y explicit, 
                y dir=both,
        ] table [
            x index=0,
            y index=2,
            y error plus expr=\thisrow{DAggerSE},
            y error minus expr=\thisrow{DAggerSE},
        ] {departs_per_meter.txt};
        
        \addplot+[blue] [
            mark=square*,
            mark options={blue},
            error bars/.cd,
                y explicit, 
                y dir=both,
        ] table [
            x index=0,
            y index=3,
            y error plus expr=\thisrow{HGDAggerSE},
            y error minus expr=\thisrow{HGDAggerSE},
        ] {departs_per_meter.txt};
        
        \addlegendentry{Behavioral Cloning}
        \addlegendentry{DAgger}
        \addlegendentry{HG-DAgger}
        
    \end{axis}
\end{tikzpicture}
    \caption{\small Mean road departure rate per meter over training epochs. Error bars represent standard deviation.}
    \label{fig:lc_road_departs}
    \vspace{-10pt}
\end{figure}
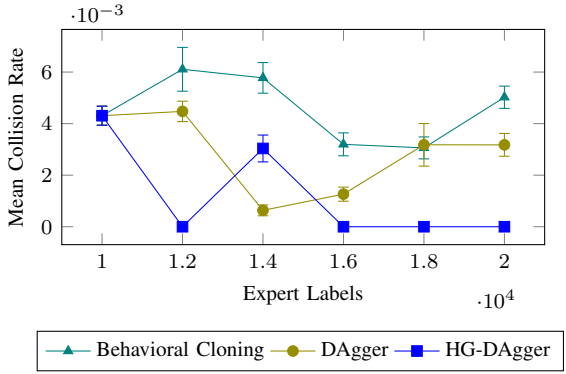
\begin{figure}[t!]
    \centering
       \begin{tikzpicture}[ trim axis right]
        \begin{axis}[
        width=0.45*\textwidth,
        height=0.25*\textwidth,
        xlabel=Expert Labels,
        ylabel=Mean Collision Rate,
        font=\footnotesize,
        legend style={
        legend columns = -1,
        at={(1.04,-0.4)},
        font=\footnotesize}
        ]
            \addplot+[teal] [
                mark=triangle*,
                mark options={teal},
                error bars/.cd,
                    y explicit,
                    y dir=both,
            ] table [
                x index=0,
                y index=1,
                y error plus expr=\thisrow{BC_SE},
                y error minus expr=\thisrow{BC_SE},
            ] {collisions_per_meter.txt};
            
            \addplot+[olive] [
                mark=*,
                mark options={olive},
                error bars/.cd,
                    y explicit, 
                    y dir=both,
            ] table [
                x index=0,
                y index=2,
                y error plus expr=\thisrow{DAgger_SE},
                y error minus expr=\thisrow{DAgger_SE},
            ] {collisions_per_meter.txt};
            
            \addplot+[blue] [
                mark=square*,
                mark options={blue},
                error bars/.cd,
                    y explicit, 
                    y dir=both
            ] table [
                x index=0,
                y index=3,
                y error plus expr=\thisrow{HG_DAgger_SE},
                y error minus expr=\thisrow{HG_DAgger_SE},
            ] {collisions_per_meter.txt};
            
            \addlegendentry{Behavioral Cloning}
            \addlegendentry{DAgger}
            \addlegendentry{HG-DAgger}
        \end{axis}
        \end{tikzpicture}
  \caption{\small Mean collision rate per meter over training epochs. Error bars represent standard deviation. 
}
    \label{fig:lc_collision_rates}
\end{figure}

HG-\textsc{DAgger} demonstrates faster and more stable learning than \textsc{DAgger} and behavioral cloning as measured by road departure rate and collision rate. All rates are per meter. 
One interesting feature of the learning curves for \textsc{DAgger} is the instability demonstrated in later epochs. As the parameter $\beta$ is decayed over training epochs, the novice policy is given control a larger percentage of the time. We hypothesize that in this situation, perceived actuator lag begins to affect the labels provided by the human expert. The observed instabilities may be a result of a concomitant deterioration in the quality of collected expert action labels.

\subsection{Simulation: Safety and Risk Evaluation}
\label{subsec:sim_safety}
We evaluated the adequacy of novice doubt as a risk metric as well as the significance of the learned doubt threshold $\tau$ by examining the novice's performance when initialized inside and outside of the estimated permissible set $\hat{\mathcal{P}}$. The set $\hat{\mathcal{P}}$ is an approximation to the true permitted set $\mathcal{P}$ derived using the doubt approximation of risk and the learned threshold $\tau$:
\begin{align}
    \hat{\mathcal{P}}
    &=
    \{
    x_t \mid
    d_N(\mathcal{O}(x_t)) \leq \tau
    \}
\end{align}
where $d_N(o_t)$ is the novice's doubt given observation $o_T$. The complement of  $\hat{\mathcal{P}}$, the estimated unsafe set, is $\hat{\mathcal{P}}'$.


To perform this experiment, we defined a set $S$ of conservative initializations. Given the ego vehicle's pose $(x,y,\theta)$, its speed $s$,
and the distance to the nearest leading obstacle in each lane $(d_l,d_r)$, we define $S$ as:
\begin{subequations}
\begin{align}
    S = \{(x,y,\theta,s) \mid & y \in [-6,6] \text{ meters},\\
        &\theta \in [-15,15] \text{ degrees},\\
        & s \in [4,5] \text{ \si{\meter\per\second}},\\
        &\textrm{max}(d_l,d_r) < 8 \text{ meters}
    \}
\end{align}
\end{subequations}
One group of initializations was sampled uniformly from  $\hat{\mathcal{P}} \cap S$ and 
and another group from $\hat{\mathcal{P}}' \cap S$. 
Using these set intersections rather than sampling uniformly over the permissible set and its complement ensures that the states sampled from $\hat{\mathcal{P}}'$ are not unrealistically dangerous states, which would be unlikely to be encountered in a real rollout.
Additionally, sampling from this set shows that our method is capable of distinguishing between similar regions of the state space on the basis of the novice policy's estimated risk within those regions. Performance was evaluated using collision rate and road departure rate.

\begin{table}
\centering
\caption{\small Mean collision and road departure rates per meter, and mean road departure duration in seconds, for rollouts initialized within or outside the permissible set.} 
\label{table:sim_safety_rates}
\tabcolsep=0.11cm
\begin{tabular}{>{\centering\arraybackslash}m{.22\linewidth}>{\centering\arraybackslash}m{.22\linewidth}>{\centering\arraybackslash}m{.22\linewidth}>{\centering\arraybackslash}m{.22\linewidth}}
\toprule
Initialization & 
Collision Rate & 
Road Departure Rate & 
Departure Duration \\
\midrule
$\hat{\mathcal{P}}$ & \num{0.607e-3} & \num{0.607e-3} & \num{1.630} \\
$\hat{\mathcal{P}}'$ & \num{7.533e-3} & \num{12.092e-3} & \num{3.740} \\
\bottomrule
\end{tabular}
\end{table}

From the results in Table~\ref{table:sim_safety_rates}, we see that the novice policy performed significantly better when initialized inside of $\hat{\mathcal{P}}$: the mean collision rate was 12 times lower and the mean road departure rate was 20 times lower than when 
initialized outside of $\hat{\mathcal{P}}$. Furthermore, the average duration of those road departures that did occur when the novice had been initialized inside of $\hat{\mathcal{P}}$ was less than half of the average duration of the road departures that occurred when the novice had been initialized outside of the permitted set.

These results highlight the utility of novice doubt as a model-free risk approximator. Furthermore, they show that HG-\textsc{DAgger} learns a doubt threshold $\tau$ that can be used to distinguish similar states that are nonetheless  distinct in terms of their riskiness.

\subsection{Test Vehicle: Driving Performance}
Policies trained on the test vehicle were evaluated on a fixed set of five random obstacle configurations in the same manner as were the policies trained in simulation. Quantitative results from these tests can be seen in Table~\ref{table:on-vehicle}. The novice trained with \textsc{HG-DAgger} had the fewest collisions and road departures of the three methods. Furthermore, the steering angle distribution induced by the \textsc{HG-DAgger} policy was 21.1\% closer to the human driving data, as measured by Bhattacharyya distance~\cite{bhattacharyya1943}, than was the distribution induced by \textsc{DAgger}, indicating more human-like behavior. However, the limited amount of on-vehicle test data limits the statistical significance of these results, and they should be interpreted primarily as a heuristic that our method could be a good candidate for further real-world evaluation.

\begin{figure}[h]
\begin{tabular}{ c }
    \includegraphics[width=.45\textwidth]{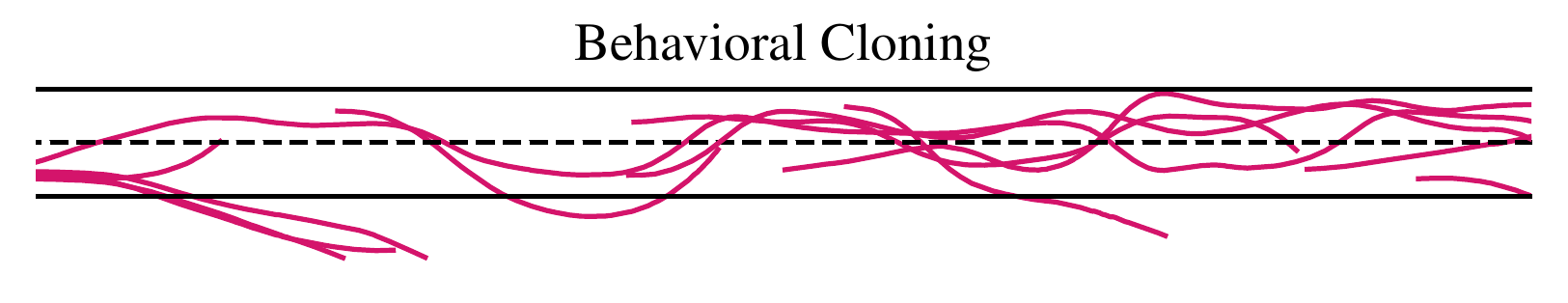} \\
    \includegraphics[width=.45\textwidth]{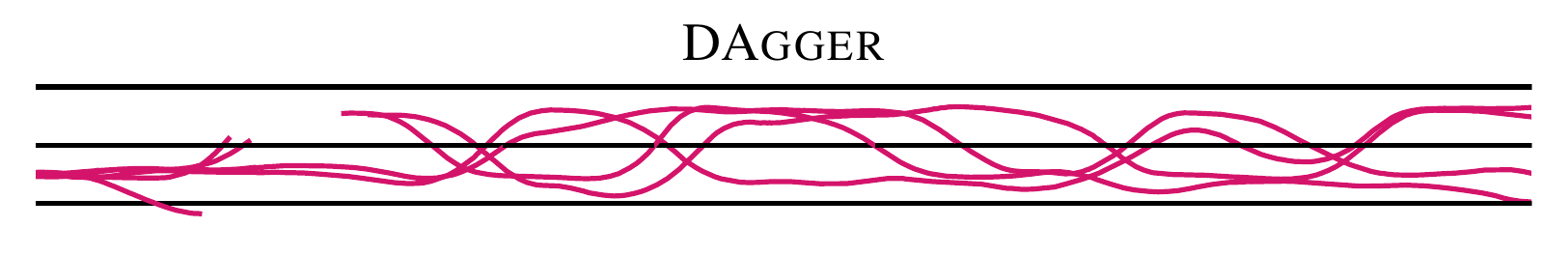} \\
    \includegraphics[width=.45\textwidth]{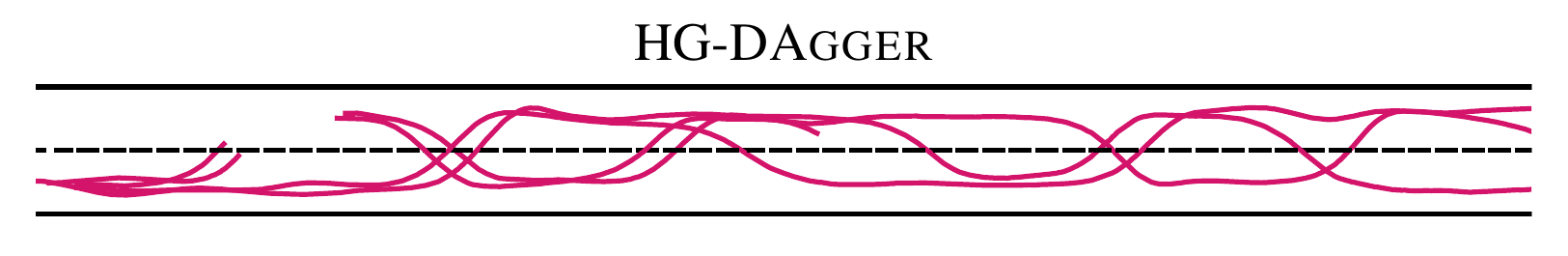}\\
    \includegraphics[width=.45\textwidth]{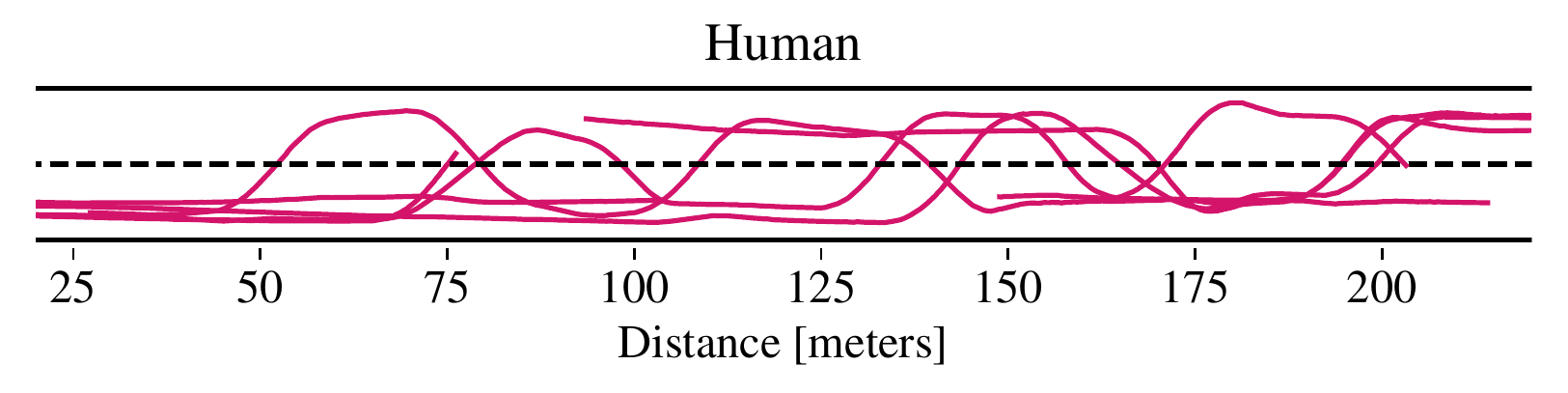} \\
\end{tabular}
\caption{\small Trajectory plots of on-vehicle test data.}
\label{fig:on-vehicle-trajs}
\vspace{-10pt}
\end{figure}

The test trajectories themselves are visualized in Fig.~\ref{fig:on-vehicle-trajs}.\footnote{Discontinuities in Fig.~\ref{fig:on-vehicle-trajs} represent points where testing had to be halted due to pedestrians or other vehicles entering the test area.} The \textsc{HG-DAgger} novice's trajectories appear qualitatively superior to those of \textsc{DAgger} and behavioral cloning, as the \textsc{HG-DAgger} policy maintains a safer distance from the edge of the road than the \textsc{DAgger} policy and does not deviate from the roadway, like the behavioral cloning policy. 

\begin{table*}[t]
\centering
\caption{Summary of on-vehicle test data. Totals are for the first 5,000 samples collected.}
\label{table:on-vehicle}
\begin{tabular}{lccccc}
\toprule
& \# Collisions & Collisions Rate & \# Road Departures & Road Departure Rate & Bhattacharyya Metric \\ \midrule
Behavioral Cloning   & \num{1}                & \num{0.973e-3}            & \num{6}                     & \num{5.837e-3}                  & \num{0.1173}                                  \\ 
\textsc{DAgger}     & \num{1}                & \num{1.020e-3}             & \num{1}                     & \num{1.020e-3}                  & \num{0.1057}                                  \\
Human-Gated \textsc{DAgger} & \textbf{\num{0}}       & \textbf{\num{0.0}}         & \textbf{\num{0}}            & \textbf{\num{0.0}}              & \textbf{\num{0.0834}}                        \\ 
\bottomrule
\end{tabular}
\vspace{-10pt}
\end{table*}


\subsection{Test Vehicle: Safety and Risk Evaluation}

A good doubt threshold $\tau$ should be low enough to exclude dangerous regions from the estimated permitted set $\hat{\mathcal{P}}$, also high enough so that it does not exclude safe regions from $\hat{\mathcal{P}}$. Therefore, to complement the results described in Section~\ref{subsec:sim_safety}, and to further validate the utility of the doubt metric and the associated threshold $\tau$ learned by HG-\textsc{DAgger}, we examined the correspondence between $\hat{\mathcal{P}}$ and free space, and between $\hat{\mathcal{P}}'$ and occupied space. Occupied space in this case corresponds to all points off of the road as well as on-road points that fall within any of the obstacle vehicles.\footnote{We note that free space and occupied space are only an approximation of safe and dangerous regions, since there exist points in free space that still fall within the region of inevitable collision.} 

We evaluate this correspondence by discretizing the workspace and using binary classification performance metrics on a pixelwise basis. Individual pixels were assigned to $\hat{\mathcal{P}}$ or $\hat{\mathcal{P}}'$ by sampling novice doubt along constant curvature trajectories and then performing linear interpolation.



\begin{figure}[!t]
\centering
\begin{minipage}{0.48\textwidth}
\begin{tabular*}{0.74\textwidth}{@{\extracolsep{\fill}}ccc@{}}
\hspace{35pt} $3\tau$ & $\tau$ & $\frac{1}{3}\tau$
\end{tabular*}
\end{minipage}
\begin{minipage}{0.48\textwidth}
\includegraphics[scale=0.25]{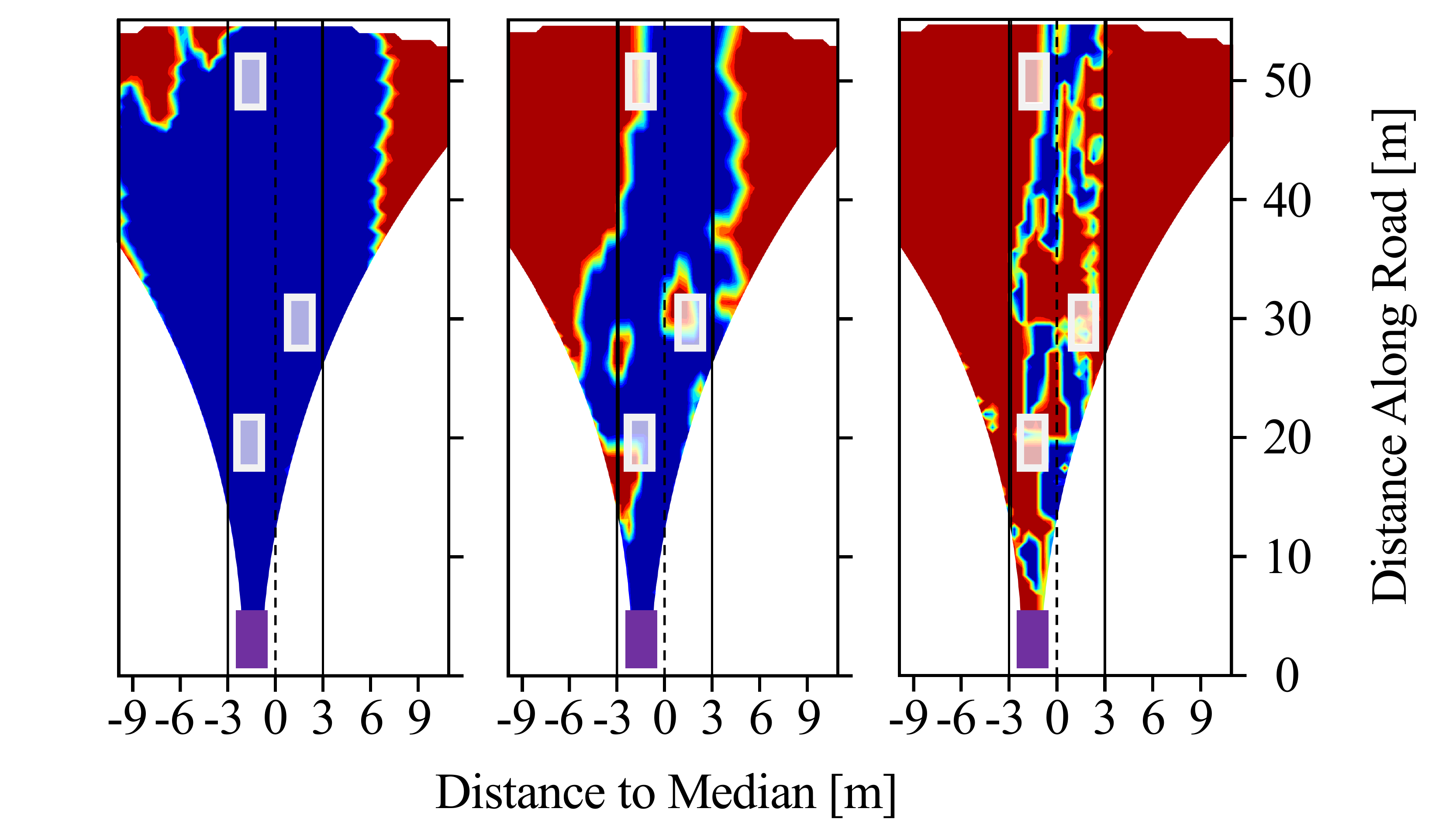}
    \caption{\small Risk maps generated for a policy trained on the test vehicle. The center map was generated using the variance threshold $\tau$ learned from human interventions. 
    The purple box represents the ego vehicle, and the white boxes represent other vehicles. Blue is safe and red is unsafe.}
    \label{fig:risk_maps}
\end{minipage}
\vspace{-5pt}
\end{figure}

Figure~\ref{fig:risk_maps} is a visualization of three risk maps generated by this process, for a single obstacle configuration, and for the doubt threshold learned by the algorithm (center), compared to two other thresholds. 
Figure~\ref{fig:risk_maps} demonstrates how the learned value of $\tau$ is meaningful: the map generated with $\tau$ provides a good approximate characterization of the riskiness of different parts of the workspace.
Increasing $\tau$ by a relatively small constant factor, however, causes dangerous regions to be inaccurately characterized as safe, while decreasing $\tau$ creates the opposite problem of an overly-conservative characterization of risk.

\begin{figure}
    \centering
       \begin{tikzpicture}[trim axis left, trim axis right]
        \begin{axis}[
        width=0.45*\textwidth,
        height=0.32*\textwidth,
        xlabel=Doubt Threshold,
        ylabel=Classification Performance,
        font=\footnotesize,
        legend style={
        legend columns = 2,
        at={(1.1,-0.25)},
        font=\footnotesize
        }
        ]
            
            \addplot+[mark options={scale=0.5}] [
                error bars/.cd,
                    y explicit,
                    y dir=both,
            ] table [
                x index=0,
                y index=2,
            ] {classification_perf.txt};
            
            \addplot+[mark options={scale=0.5}] [
                error bars/.cd,
                    y explicit,
                    y dir=both,
            ] table [
                x index=0,
                y index=3,
            ] {classification_perf.txt};
            
            \addplot+[teal,mark options={scale=0.5}] [
                error bars/.cd,
                    y explicit,
                    y dir=both,
            ] table [
                x index=0,
                y index=4,
            ] {classification_perf.txt};
            
            \addplot+[olive,mark options={scale=0.5}] [
                error bars/.cd,
                    y explicit,
                    y dir=both,
            ] table [
                x index=0,
                y index=5,
            ] {classification_perf.txt};
            
            \addplot+[orange,mark options={scale=0.5}] [
                error bars/.cd,
                    y explicit,
                    y dir=both,
            ] table [
                x index=0,
                y index=6,
            ] {classification_perf.txt};
            
           \addplot[mark=none, very thick, black, dashed] coordinates {(0.261,0) (0.261,1)};
            
            \addlegendentry{Micro-Average F1 Score}
            \addlegendentry{Average F1 Score}
            \addlegendentry{Balanced Accuracy}
            \addlegendentry{Free Space F1 Score}
            \addlegendentry{Occupied Space F1 Score}
            \addlegendentry{Learned Doubt Threshold}
            
        \end{axis}
        \end{tikzpicture}
  \caption{\small Performance on the pixelwise free vs. occupied space classification task as a function of the doubt threshold used.} 
    \label{fig:classification_performance}
    \vspace{-20pt}
\end{figure}
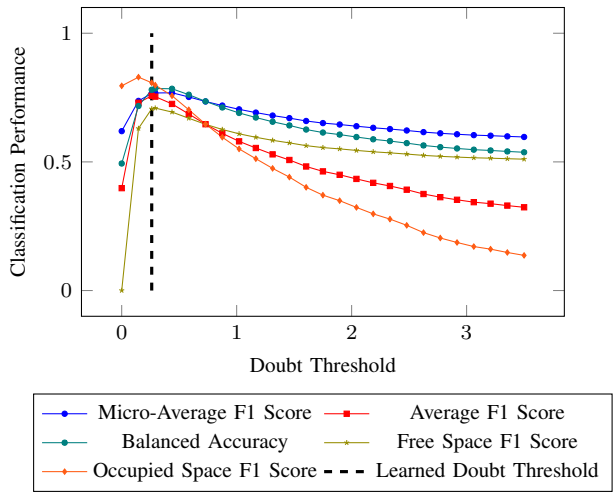

Figure~\ref{fig:classification_performance} demonstrates that these results are not limited to the single obstacle configuration seen in Figure~\ref{fig:risk_maps}. The chart shows various performance metrics for the pixel-wise free space vs. occupied space classification task as functions of the doubt threshold used. Each choice of threshold was evaluated on 40 randomly generated obstacle configurations. The chart shows that the threshold learned from HG-\textsc{DAgger} is near-optimal for all performance metrics examined.

\section{Conlcusion}
\label{sec:disc}


While interactive imitation learning algorithms like $\textsc{DAgger}$ address some of the limitations of behavioral cloning, it is not straightforward to apply such methods to the task of learning from human experts. By limiting the control authority of the human expert, these methods can degrade the quality of collected action labels and can compromise the safety of the combined expert-novice system.

In response to these challenges, we have presented HG-\textsc{DAgger}, a \textsc{DAgger}-variant that enables more effective learning in the human-in-the-loop context. HG-\textsc{DAgger} learns both a novice policy and a safety threshold for a doubt metric that can be used to estimate risk in different regions of the state space at test-time. We demonstrated the efficacy of our method and of the learned risk threshold in simulated and real-world autonomous driving experiments, showing improved sample efficiency and increased training stability relative to \textsc{DAgger} and behavioral cloning.

Future work will involve the use of doubt-based risk metrics as inputs to an automated gating mechanism to switch between sub-policies in a hierarchical controller. We also plan to investigate more sophisticated ways of estimating model uncertainty 
and linking it with execution risk.

\section*{Acknowledgments}
We thank Kunal Menda for inspiring this work and Xiaobai Ma for his help interfacing with the test vehicle.


\bibliographystyle{IEEEtran}
\bibliography{BibFile}

\begin{thebibliography}{10}
\providecommand{\url}[1]{#1}
\csname url@rmstyle\endcsname
\providecommand{\newblock}{\relax}
\providecommand{\bibinfo}[2]{#2}
\providecommand\BIBentrySTDinterwordspacing{\spaceskip=0pt\relax}
\providecommand\BIBentryALTinterwordstretchfactor{4}
\providecommand\BIBentryALTinterwordspacing{\spaceskip=\fontdimen2\font plus
\BIBentryALTinterwordstretchfactor\fontdimen3\font minus
  \fontdimen4\font\relax}
\providecommand\BIBforeignlanguage[2]{{%
\expandafter\ifx\csname l@#1\endcsname\relax
\typeout{** WARNING: IEEEtran.bst: No hyphenation pattern has been}%
\typeout{** loaded for the language `#1'. Using the pattern for}%
\typeout{** the default language instead.}%
\else
\language=\csname l@#1\endcsname
\fi
#2}}

\bibitem{price2003accelerating}
B.~Price and C.~Boutilier, ``Accelerating reinforcement learning through
  implicit imitation,'' \emph{Journal of Artificial Intelligence Research},
  vol.~19, pp. 569--629, 2003.

\bibitem{Kober2010}
J.~Kober and J.~Peters, ``Imitation and reinforcement learning,'' \emph{IEEE
  Robotics Automation Magazine}, vol.~17, no.~2, pp. 55--62, June 2010.

\bibitem{ross2010efficient}
S.~Ross and D.~Bagnell, ``Efficient reductions for imitation learning,'' in
  \emph{International Conference on Artificial Intelligence and Statistics},
  2010, pp. 661--668.

\bibitem{laskey2017comparing}
M.~Laskey, C.~Chuck, J.~Lee, J.~Mahler, S.~Krishnan, K.~Jamieson, A.~Dragan,
  and K.~Goldberg, ``Comparing human-centric and robot-centric sampling for
  robot deep learning from demonstrations,'' in \emph{IEEE International
  Conference on Robotics and Automation (ICRA)}, 2017, pp. 358--365.

\bibitem{goil2013using}
A.~Goil, M.~Derry, and B.~D. Argall, ``Using machine learning to blend human
  and robot controls for assisted wheelchair navigation,'' in \emph{IEEE
  International Conference on Rehabilitation Robotics (ICORR)}, 2013.

\bibitem{kim1992force}
W.~S. Kim, B.~Hannaford, and A.~Fejczy, ``Force-reflection and shared compliant
  control in operating telemanipulators with time delay,'' \emph{IEEE
  Transactions on Robotics and Automation}, vol.~8, no.~2, pp. 176--185, 1992.

\bibitem{shia2014semiautonomous}
V.~A. Shia, Y.~Gao, R.~Vasudevan, K.~{Driggs-Campbell}, T.~Lin, F.~Borrelli,
  and R.~Bajcsy, ``Semiautonomous vehicular control using driver modeling,''
  \emph{IEEE Transactions on Intelligent Transportation Systems}, vol.~15,
  no.~6, pp. 2696--2709, 2014.

\bibitem{branicky1998multiple}
M.~S. Branicky, ``Multiple {Lyapunov} functions and other analysis tools for
  switched and hybrid systems,'' \emph{IEEE Transactions on Automatic Control},
  vol.~43, no.~4, pp. 475--482, 1998.

\bibitem{mcruer1995pilot}
D.~T. McRuer, ``Pilot-induced oscillations and human dynamic behavior,''
  \emph{NASA Technical Report}, 1995.

\bibitem{national1997aviation}
D.~McRuer, \emph{Aviation safety and pilot control: Understanding and
  preventing unfavorable pilot-vehicle interactions}.\hskip 1em plus 0.5em
  minus 0.4em\relax National Academies Press, 1997.

\bibitem{zhang2016query}
J.~Zhang and K.~Cho, ``Query-efficient imitation learning for end-to-end
  autonomous driving,'' \emph{arXiv}, no. 1605.06450, 2016.

\bibitem{laskey2016shiv}
M.~Laskey, S.~Staszak, W.~Y.-S. Hsieh, J.~Mahler, \emph{et~al.}, ``Shiv:
  Reducing supervisor burden in {DA}gger using support vectors for efficient
  learning from demonstrations in high dimensional state spaces,'' in
  \emph{IEEE International Conference on Robotics and Automation (ICRA)}, 2016,
  pp. 462--469.

\bibitem{kim2013maximum}
B.~Kim and J.~Pineau, ``Maximum mean discrepancy imitation learning,'' in
  \emph{Robotics: Science and Systems}, 2013.

\bibitem{menda2018ensembledagger}
K.~Menda, K.~Driggs-Campbell, and M.~J. Kochenderfer,
  ``{\textsc{EnsembleDAgger}}: A {B}ayesian approach to safe imitation
  learning,'' \emph{arXiv}, no. 1807.08364, 2018.

\bibitem{driggs2015improved}
K.~Driggs-Campbell, V.~Shia, and R.~Bajcsy, ``Improved driver modeling for
  human-in-the-loop vehicular control,'' in \emph{IEEE International Conference
  on Robotics and Automation (ICRA)}, 2015.

\bibitem{govindarajan2017data}
V.~Govindarajan, K.~Driggs-Campbell, and R.~Bajcsy, ``Data-driven reachability
  analysis for human-in-the-loop systems,'' in \emph{IEEE Conference on
  Decision and Control (CDC)}, 2017, pp. 2617--2622.

\bibitem{lakshminarayanan2017simple}
B.~Lakshminarayanan, A.~Pritzel, and C.~Blundell, ``Simple and scalable
  predictive uncertainty estimation using deep ensembles,'' in \emph{Advances
  in Neural Information Processing Systems (NIPS)}, 2017, pp. 6405--6416.

\bibitem{chernova2009}
S.~Chernova and M.~Veloso, ``Interactive policy learning through
  confidence-based autonomy,'' in \emph{Journal of Artificial Intelligence
  Research (JAIR)}, 2009, pp. 1--25.

\bibitem{heheCoaching}
H.~He, H.~Daum\'e~III, and J.~Eisner, ``Imitation learning by coaching,'' in
  \emph{Thirty-second Conference on Neural Information Processing System},
  2012.

\bibitem{pomerleau1989alvinn}
D.~A. Pomerleau, ``{ALVINN}: An autonomous land vehicle in a neural network,''
  in \emph{Advances in Neural Information Processing Systems}, 1989, pp.
  305--313.

\bibitem{bojarski2016end}
M.~Bojarski, D.~Del~Testa, D.~Dworakowski, B.~Firner, B.~Flepp, P.~Goyal, L.~D.
  Jackel, M.~Monfort, U.~Muller, J.~Zhang, X.~Zhang, J.~Zhao, and K.~Zieba,
  ``End to end learning for self-driving cars,'' \emph{arXiv}, no. 1604.07316,
  2016.

\bibitem{pan2017agile}
Y.~Pan, C.-A. Cheng, K.~Saigol, K.~Lee, X.~Yan, E.~Theodorou, and B.~Boots,
  ``Agile off-road autonomous driving using end-to-end deep imitation
  learning,'' \emph{arXiv}, no. 1709.07174, 2017.

\bibitem{bhattacharyya1943}
A.~Bhattacharyya, ``On a measure of divergence between two statistical
  populations defined by their probability distributions,'' \emph{Bulletin of
  the Calcutta Mathematical Society}, vol.~35, pp. 99--109, 1943.

\end{thebibliography}

\end{document}